\documentclass{article}

\usepackage{microtype}
\usepackage{graphicx}
\usepackage{subfigure}
\usepackage{booktabs} 
\usepackage{multirow}

\usepackage{hyperref}

\usepackage[accepted]{icml2019}

\icmltitlerunning{How much real data do we actually need:
   	 Analyzing object detection performance using synthetic and real data}

\begin{document}

\twocolumn[
\icmltitle{How much real data do we actually need:\\
   	 Analyzing object detection performance using synthetic and real data}



\icmlsetsymbol{equal}{*}

\begin{icmlauthorlist}
\icmlauthor{Farzan Erlik Nowruzi}{se,uo}
\icmlauthor{Prince Kapoor}{se}
\icmlauthor{Dhanvin Kolhatkar}{se,uo}
\icmlauthor{Fahed Al Hassanat}{se,uo}
\icmlauthor{Robert Laganiere}{se,uo}
\icmlauthor{Julien Rebut}{va}
\end{icmlauthorlist}

\icmlaffiliation{se}{Sensor Cortek Inc.}
\icmlaffiliation{uo}{University of Ottawa}
\icmlaffiliation{va}{Valeo AI}

\icmlcorrespondingauthor{Farzan Erlik Nowruzi}{erlik@sensorcortek.ai}

\icmlkeywords{Object Detection, Autonomous Driving, Deep Learning, Computer Vision, Machine Learning}

\vskip 0.3in
]



\printAffiliationsAndNotice{}  

\begin{abstract}
	In recent years, deep learning models have resulted in a huge amount of progress in various areas, including computer vision. By nature, the supervised training of deep models requires a large amount of data to be available. This ideal case is usually not tractable as the data annotation is a tremendously exhausting and costly task to perform. An alternative is to use synthetic data. In this paper, we take a comprehensive look into the effects of replacing real data with synthetic data. We further analyze the effects of having a limited amount of real data. We use multiple synthetic and real datasets along with a simulation tool to create large amounts of cheaply annotated synthetic data. We analyze the domain similarity of each of these datasets. We provide insights about designing a methodological procedure for training deep networks using these datasets.
\end{abstract}
\section{Introduction}
\textit{Deep learning} models have revolutionized the field of computer vision. These models have been applied on various tasks such as object detection \cite{SpeedAccuracy, Wang2018, Simon2019}, scene segmentation \cite{FCN, Deeplab, PSPNet}, crowd analysis \cite{KangMC17}, and autonomous driving \cite{NvidiaCars, multinet}.

Autonomous driving is one of the hottest fields that is benefiting from the availability of such models. Deep models are commonly used for vehicle detection \cite{VehicleLidar}, pedestrain detection \cite{Xu_2017, Zhang_2016}, open road segementation \cite{Chen_2018, multinet}, and end to end driving \cite{NvidiaCars, codevilla2018end}.

The idea of deep neural networks is not a new concept.
The major success of these models can be attributed to the availability of required computational power to perform the huge amount of calculations, and to the availability of large datasets to learn the transformation functions.

Many studies \cite{RastegariORF16, ENet, SpeedAccuracy} have been undertaken to address the computational costs associated with training and inference on deep models. Regularly, new datasets \cite{ImageNet, MSCOCO, cityscapes, kitti, bdd100k, P4B, nuscenes2019} are introduced to provide the required diversity of information for the models to learn an specific task.
Collection of raw data, annotation and verification of these datasets is a very expensive and time-consuming task. As such, researchers have been developping various techniques to overcome this issue and introduce cost saving measures to build a high quality dataset.

Domain adaptation \cite{Sankaranarayanan2017, Carlson2018, Hoffman2017} techniques try to adapt the network from source to target domain by comparing feature maps of each domain \cite{Hoffman2016}, using Generative Adversarial Networks \cite{GAN} to alter the features of target domain such that they match the features of the source domain \cite{Sankaranarayanan2017, Wu2018, Zhang2018}, or modifying the style of an image in the target domain to match the source domain for feature extraction \cite{nuscenes2019, Zhang2018}. Few-shot learning \cite{Qiao2017, Kang2018, Sung2017} is another field that deals with the difficulties of annotating a large amount of data required for deep learning.
While these are promising approaches, they are still not satisfactory in dealing with the major domain shift in the underlying distribution of the datasets \cite{rao2018deep}.

We cannot underestimate the tremendous importance of having a large amount of annotated data in order to train models that could generalize to new test cases. A promising method in this direction is to create simulated data that is well capable of imitating the statistics of the real data. In this study, we explore multiple avenues in this context by focusing on various synthetic datasets and their interaction with multiple real datasets. Our goal is to identify a procedure that addresses the domain shift and dataset completeness challenges. Our work follows a similar approach as \cite{Mayer2018,7D} with a focus on autonomous driving and object detection in order to find a best practice procedure for training deep models. More formally, we define our contributions as follows:
\begin{itemize}
    \item An analysis about the adverse effects of reduction in dataset size on the performance of the object detection models.
	\item A comprehensive metric study to show the relationship between underlying distributions of various datasets.
	\item Mixed training on large synthetic datasets and small real datasets.
	\item Fine-tuning models trained on large synthetic datasets with a small set of real data.
	\item Evaluating the effect of training on multiple synthetic datasets at once, on the fine-tuning performance on small sized real dataset.
\end{itemize}

It is worth to mention that the quality of a dataset is not only measured by its size. There are factors such as diversity, completeness, appearance, object occurrence distribution and other factors that impact the effectiveness of a dataset. In this paper, we mainly focus on addressing the quantity aspect of real datasets, which have a higher collection and annotation cost.

Finally, from a deployment point of view, we consider an architecture that is deployable on embedded platforms. For this purpose, we consider the MobileNet \cite{mobilenet} feature extractor as a backbone to the single shot detector (SSD) \cite{SSD}. The SSD-MobileNet combination is shown to be a fast and commonly used model in object detection tasks \cite{mobilenet, SpeedAccuracy, ssdmobilenet1}.


\begin{figure*}[!ht]
\vskip 0.2in
\begin{center}
   (a) \includegraphics[width=0.3\linewidth]{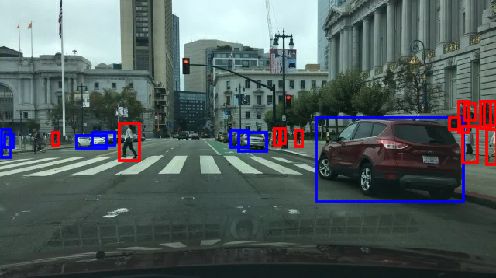}
   (b) \includegraphics[width=0.3\linewidth]{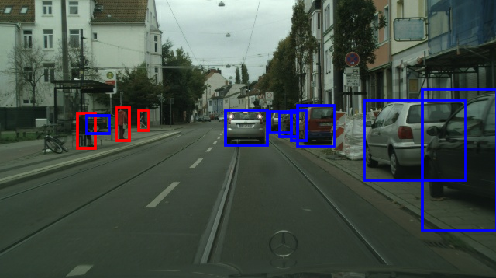}
   (c) \includegraphics[width=0.3\linewidth]{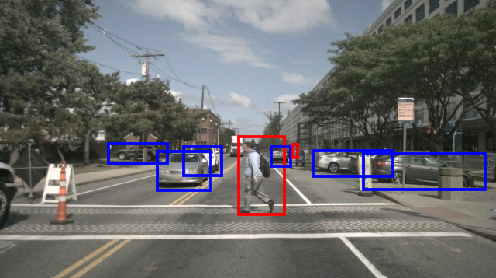}
   (d) \includegraphics[width=0.3\linewidth]{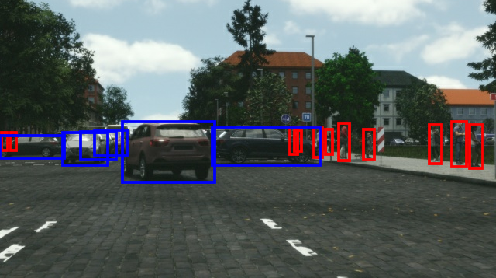}
   (e) \includegraphics[width=0.3\linewidth]{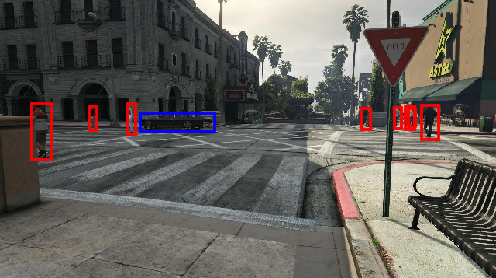}
   (f) \includegraphics[width=0.3\linewidth]{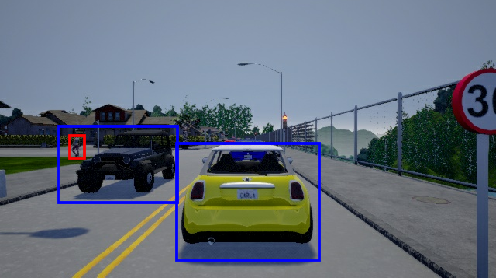}
\end{center}
\vskip -0.1in
\caption{Sample images from real and synthetic datasets.(a) BDD~\cite{bdd100k}, (b) KC~\cite{kitti}\cite{cityscapes}, (c) NS~\cite{nuscenes2019}, (d) 7D~\cite{7D}, (e) P4B~\cite{P4B}, (f) CARLA~\cite{CARLA}.}
\label{fig:reduction}
\end{figure*}

\begin{table*}[h]
\caption{List of datasets and their statistics used in our experiments grouped for both car and person classes. Statistics, in order, include average number of object class per image, percentage of small, medium and large bounding boxes for each class, and average aspect ratio (\textit{AR}) for the objects. The final column (\textit{car/person}) shows the ratio of number of total car annotations to the number of person annotations for each dataset. Objects with height smaller than 4\% of the image height are removed. Cars with area less than 1\% of image area are grouped as Small, remaining cars with area less than 2.5\% are grouped as Medium, and the rest are considered as Large. For person class, Small objects have area less than 0.2\%, Medium group contains remaining bounding boxes with area less than 1\%, and the rest of them are in the Large group.}
\label{tab:datasets}
\vskip 0.15in
\begin{center}
\begin{small}
\begin{sc}
\resizebox{2.05\columnwidth}{!}{%
\begin{tabular}{p{2.5cm}c|ccccc|ccccc|ccc}
\toprule
Dataset & Type  & \textbf{car} & Small(\%) & Medium(\%) & Large(\%) & AR & \textbf{person} & Small(\%) & Medium(\%) & Large(\%) & AR & \textbf{car/person} \\
\midrule
BDD~\cite{bdd100k}& Real &6.09 &67.4&15.5&17.2&1.30 &1.38 &60.6&32.4&7.0& 0.41 & 4.41 \\
KC~\cite{kitti}\cite{cityscapes} & Real &3.80 &49.7&20.7&29.7&1.58 &0.93 &30.5&35.4&27.1& 0.39 & 4.09 \\
NS~\cite{nuscenes2019}& Real &2.64 &60.6&21.3&18.2&1.77 &1.34 &50.3&37.7&12.0& 0.52& 1.97 \\
\midrule
7D~\cite{7D}& Synt &3.40 &46.3&25.2&28.5& 1.73 &10.99&46.4&40.4&13.2& 0.34 & 0.31 \\
P4B~\cite{P4B}& Synt &1.45 &60.8&17.1&22.1&1.93 &1.09 &69.2&21.9&8.9& 0.38 & 1.33 \\
CARLA~\cite{CARLA}& Synt &1.91 &49.7&21.1&29.2&2.02 &1.58 &58.2&29.1&12.8& 0.53& 1.21 \\
\bottomrule
\end{tabular}
}
\end{sc}
\end{small}
\end{center}
\vskip -0.1in
\end{table*}

\section{Literature Review}

The necessity of large amounts of annotated data is a bottleneck in computer vision tasks. One way of dealing with this issue is to use cheap synthetically generated training images. However, this approach posesses an important issue; how the synthetic and real data should be used to optimize training of a model.

Synthetic data generation can be undertaken in two main ways:
\begin{itemize}
    \item Real data augmentation~\cite{Tremblay2018, Fang2018}: adding objects to existing frames, which does not require sophisticated environmental modelling nor estimating spawn probabilities.
    \item Synthetic data generation through simulation~\cite{Tsirikoglou2017, 7D, P4B}: generating entire frames with sophisticated environmental modelling, sensor noise modelling and rendering. Commercial game engines~\cite{P4B, Johnson-Roberson2017} are an alternative in this domain.
\end{itemize}

\cite{Tremblay2018} uses domain randomization for car detection by effectively abandoning photorealism in the creation of the synthetic dataset.
The main motivation of \cite{Tremblay2018} is to force the network into learning only the essential features of the task.
Results of this approach are comparable to the Virtual KITTI dataset \cite{VKITTI} which closely resembles the KITTI dataset \cite{kitti}. It is also shown that using real images in the fine tuning stage, the accuracy of the models are consistently improved.
One issue with the Virtual KITTI dataset is its limited sample size of 2500 images. This could result in worse performance than the larger datasets. Further, analyzed networks are very large and very challenging to use on embedded platforms.


In contrast to \cite{Tremblay2018}, \cite{AbuAlhaija2018} uses real scenes and augments them with the synthetic objects. This way, the complexity associated with the creation of 3D environment models and spawn probabilities are avoided. It is shown that after augmenting the dataset with certain number of images a saturation point is reached. Cluttering the images with too many objects also reduces the model performance.
Another interesting finding of this work is about the significance of proper background and camera noise modeling.

A fast synthetic data generation approach is proposed in \cite{Johnson-Roberson2017}. \textit{Grand Theft Auto} (GTA) game is used to collect this dataset. It is argued that current deep vehicle detection models are usually trained on a specific dataset and cannot generalize to other datasets. Faster-RCNN \cite{FasterRCNN} is used to be trained on a large simulated dataset and on a small real dataset \cite{cityscapes}. The lack of inclusion of fine-tuning or mixed training is a missing point from this study.
\cite{P4B} follows the same idea of using GTA to generate a synthetic dataset, however their study does not include an analysis about object detection performance.

\cite{Wu2017} employes GTA to create a synthetic LiDAR dataset and uses it in combination with KITTI \cite{kitti} to train a deep model that can outperform each individual dataset. \cite{Fang2018} introduces a framework to generate simulated 3D point clouds based on physical characteristics of the LiDAR. First, a LiDAR and a camera are used to create a real background environment dataset. This data is then augmented with synthetic 3D objects, hence reducing the burden of simulated environment model creation.

\cite{Mayer2018} provides a comprehensive analysis about using various synthetic and real datasets for training neural networks for optical flow and disparity estimation tasks. Authors point to four important findings. The first is the importance of diversity in the training set. Secondly, they show that the photo-realism of the dataset is not a significant factor for the good performance of the model. It is shown that increasing dataset complexity on schedules helps to increase the performance. And finally, modelling the camera distortion of the test set is found to be highly valuable.

Synscapes (7D)~\cite{7D} is a street scene synthetic dataset that uses an end-to-end approach to generate scenarios and high fidelity rendered images. 7D is a better dataset than P4B for testing object detection models trained on a real dataset. However, using the sole 7D dataset for training performs poorly in tests on real data, and fine-tuning it on real datasets yields a better outcome.

The findings from the literature can be concluded in few rules of thumb; more data is almost always better than less data.
Samples acquired from the test environment contribute more to the performance of the model.
Augmenting real images saturate once a specific ratio is passed.
Realistic sensor distortion models and environment distribution models have a larger effect on the final performance than the photo-realism of the samples.


\section{Datasets and Data Statistics}

To perform a comprehensive study on the requirements of having expensive real data, we use a set of real and synthetic datasets. The datasets are chosen to provide comprehensive sets of examples for camera based object detection. In this paper we are only evaluating for two classes; cars and persons, as they have a more unified definition among various datasets. The list of datasets are shown in Table \ref{tab:datasets}.


Images in each dataset are resized to have an aspect ratio of $16:9$ and the shape of $640\times370$ to keep the computation tractable. From all of the datasets, the objects that have a height smaller than $4\%$ of the image height are removed. There are few reasons for this choice. First, smaller region of interest will result in a very small feature map regions. In most cases, they will be ignored. This will result in a higher loss value for SSD in detecting objects that are not visible. Secondly, smaller objects are at further distances and they might not be very relevant for immediate consideration in the task of autonomous driving.

We are constrained to $15000$ images from each dataset, which is the maximum number of images existing in the smallest dataset. To keep data distributions similar, we only use images that are in good weather conditions.

\subsection{Real Datasets}

\paragraph{Berkeley Deep Drive (BDD)}
\cite{bdd100k} consists of 100K images with annotations for segmentation and objects. Only images captured in the \textit{\{daytime, dawn/dusk\}} and under \textit{\{'clear', 'undefined', 'partly cloudy', 'overcast'\}} weather conditions are considered for sampling.

\paragraph{Kitti-CityScapes (KC)}
KITTI~\cite{kitti} has $7481$ images with aspect ratio of 33:10. To deal with the ultra wide aspect ratio, they are divided into two 16:9 images with an overlap of $14.38\%$ in between them. This way we get a total number of $14962$, which does not reach the predefined 15K images. The Cityscapes dataset~\cite{cityscapes} is visually similar to KITTI. There are $2975$ images with instance segmentation provided in CityScapes. We use the segmentation information to create bounding boxes. We sample $2000$ from CityScapes and 13K from KITTI to created the combined Kitti-CityScapes dataset.

\paragraph{NuScenes (NS)}
\cite{nuscenes2019} is a recently published dataset that contains $1000$ driving scenarios with a duration of $20$ seconds. This dataset includes readings with a complete suite of sensors including $6$ cameras, $1$ LiDAR, $5$ radars, a GPS, and an IMU. To keep the dataset consistent with the others, we only choose the frontal view camera images.

\subsection{Synthetic Datasets}

\paragraph{Synscapes (7D)}
\cite{7D} is a street scene dataset with 25K images that mimics the properties of Cityscapes \cite{cityscapes}. We choose only the bounding boxes that have occlusion values less than $70\%$. As shown in Table \ref{tab:datasets}, Synscapes has the largest deviation from the real datasets in terms of average number of objects per image, more specifically, in the number of persons per image.

\paragraph{Playing for Benchmark (P4B)}
\cite{P4B} is a dataset consisting of video sequences with a total of 200K frames. We sample every 9 frames from each video and collect a dataset of 15k images. We drop the \textit{night} sequences to comply with the conditions of other datasets.

\paragraph{CARLA}
\cite{CARLA} is a driving simulator that contains $5$ different towns, with a large variety of cars, while limited in its number of pedestrian models. We use all the $5$ towns of CARLA and generate a uniform number of $3000$ images per town.

\section{Experiments}
Our experiments are crafted to evaluate the performance of car and person detection for the autonomous driving task using a diverse combination datasets. Our main goal is to propose a procedure for training deep models using large amounts of synthetic and limited amounts of real data.

For the model of choice, SSD-MobileNet, we use the default parameter set provided in Tensorflow object detection API \cite{SpeedAccuracy} and train the model from scratch. For all of the training sessions, we use RMSprop optimizer with initial learning rate of 0.004, decay factor of 0.95 and the decay step of 30K. In fine-tuning, we change the decay step to 5000 and adjust the decay factor to 0.8.

We report the results of our experiments in terms of average precision and recall. To calculate each, the intersection of union is set at $[0.5, 0.95]$ with step size of $0.05$, and the score threshold for detected bounding boxes is set to $[0.05, 0.95]$ with a step size of $0.05$. All the results are averaged on a per-class basis and the average precision shown in figures is the average over the two classes.

\medskip
We perform our experimental analysis in four categories. For all datasets, we create a test set consisting of $2000$ samples, and use the remaining 13K samples as their training set.

\begin{table}[!ht]
\caption{Synthetic and Real Data Ratios.   Various ratios used for the synthetic and real data for the training set.}
\label{tab:ratios}
\vskip 0.15in
\begin{center}
\begin{small}
\begin{sc}
\begin{tabular}{ccc}
\toprule
Experiment & Synthetic Ratio & Real Ratio \\
\midrule
Exp 1     & 0\%    	 & 100\% \\
Exp 2     & 90\%     & 10\% \\
Exp 3     & 95\%     & 5\% \\
Exp 4      & 97.5\%     & 2.5\% \\
\bottomrule
\end{tabular}
\end{sc}
\end{small}
\end{center}
\vskip -0.1in
\end{table}

\subsection{Dataset Reduction}
In this section, we are evaluating the effects of dataset reduction on the model's performance to set the basis for our benchmarks. For this task, only the real datasets are evaluated. We choose training dataset sizes of $100\%$, $10\%$, $5\%$ and $2.5\%$. Figure \ref{fig:reduction} shows the collective results and table \ref{tbl:reduction} shows them for individual classes.

\begin{figure}[!h]
\vskip 0.2in
\begin{center}
   \includegraphics[width=0.99\linewidth]{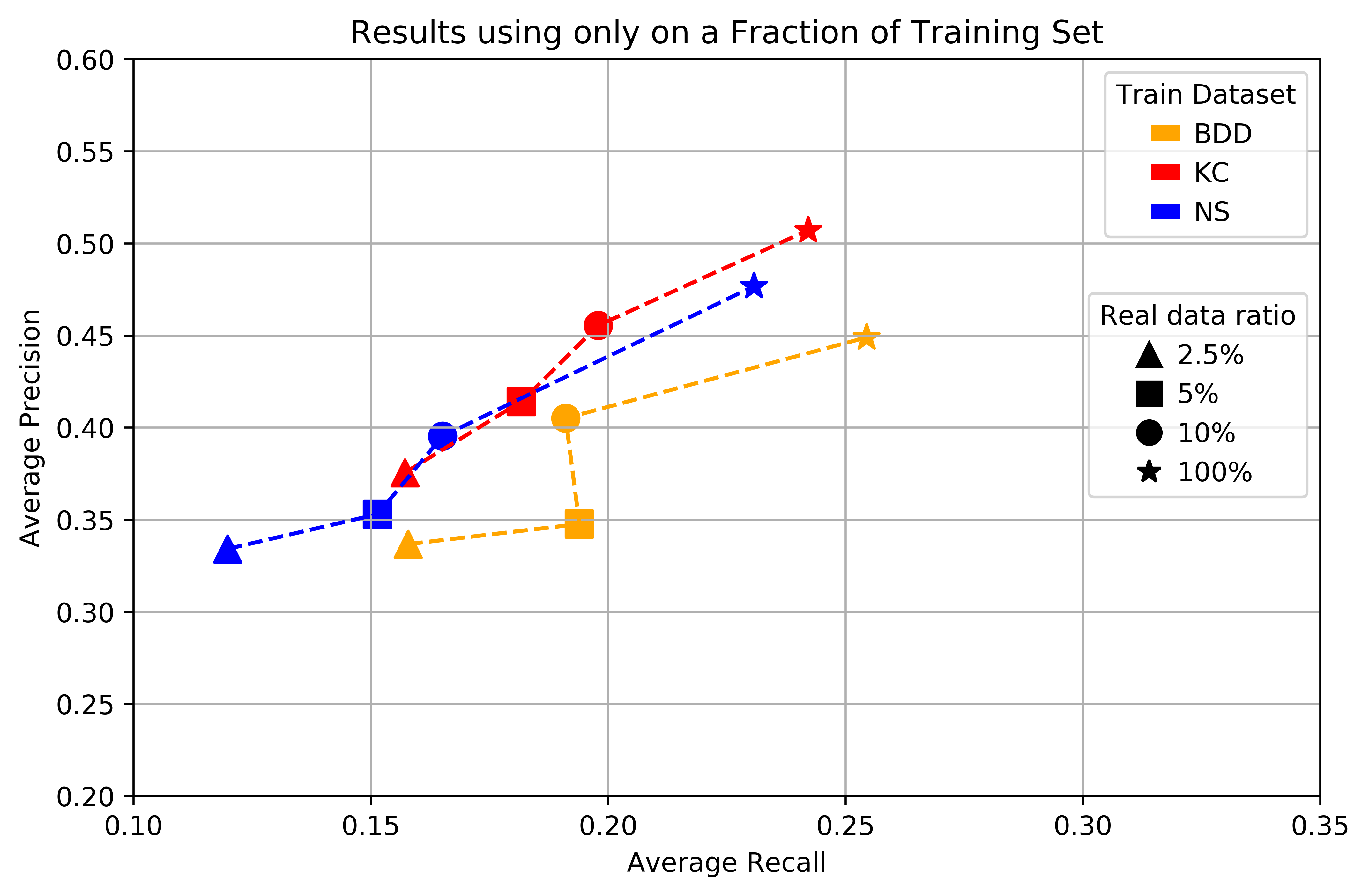}
\end{center}
\vskip -0.2in
   \caption{Effect of reducing training dataset size on the performance of the model on their corresponding test set.}
\label{fig:reduction}
\end{figure}

The general trend is that by reducing the number of real data points, we are dramatically sacrificing performance in both precision and recall terms. We also observe that, on all datasets, the relative effect of removing the first $90\%$ of data is less than the effect of removing the next $5\%$.

\begin{table}[!ht]
\caption{Results for individual classes while various training dataset sizes are used.}
\vskip 0.15in
\begin{center}
\begin{small}{}
\begin{sc}
\resizebox{.95\columnwidth}{!}{%
\begin{tabular}{lr|cc|cc}
\toprule
\multicolumn{2}{c|}{\textbf{Training Dataset}}    & \multicolumn{2}{c|}{\textbf{Person}}                  & \multicolumn{2}{c}{\textbf{Car}}                    \\
\multicolumn{1}{c}{Dataset} & \multicolumn{1}{c|}{Size} & \multicolumn{1}{c}{Precision} & \multicolumn{1}{c|}{Recall} & \multicolumn{1}{c}{Precision} & \multicolumn{1}{c}{Recall} \\
\midrule
\multirow{3}{*}{BDD}
&100\% &0.279 &0.087 &0.619 &0.422 \\
&10\%  &0.253 &0.041 &0.557 &0.341 \\
&5\%   &0.158 &0.043 &0.538 &0.344 \\
&2.5\% &0.158 &0.031 &0.515 &0.285 \\
\midrule
\multirow{3}{*}{KC}
&100\% &0.390 &0.121 &0.624 &0.364 \\
&10\%  &0.333 &0.085 &0.578 &0.311 \\
&5\%   &0.278 &0.070 &0.550 &0.293 \\
&2.5\% &0.244 &0.049 &0.507 &0.265 \\
\midrule
\multirow{3}{*}{NS}
&100\% &0.381 &0.117 &0.572 &0.345 \\
&10\%  &0.298 &0.064 &0.492 &0.266 \\
&5\%   &0.264 &0.052 &0.442 &0.251 \\
&2.5\% &0.218 &0.040 &0.450 &0.199 \\
\bottomrule
\end{tabular}
}
\label{tbl:reduction}
\end{sc}
\end{small}
\end{center}
\vskip -0.1in
\end{table}

It is also observed that precision and recall for 'person' class is significantly lower than the 'car' class. This is attributed to the vast diversity and deformability of the object shape of the person. These results are used as the baseline for the rest of the experiments in the paper.

\subsection{Dataset Similarity}
\begin{figure}[!ht]
\vskip 0.2in
\begin{center}
   \includegraphics[width=0.99\linewidth]{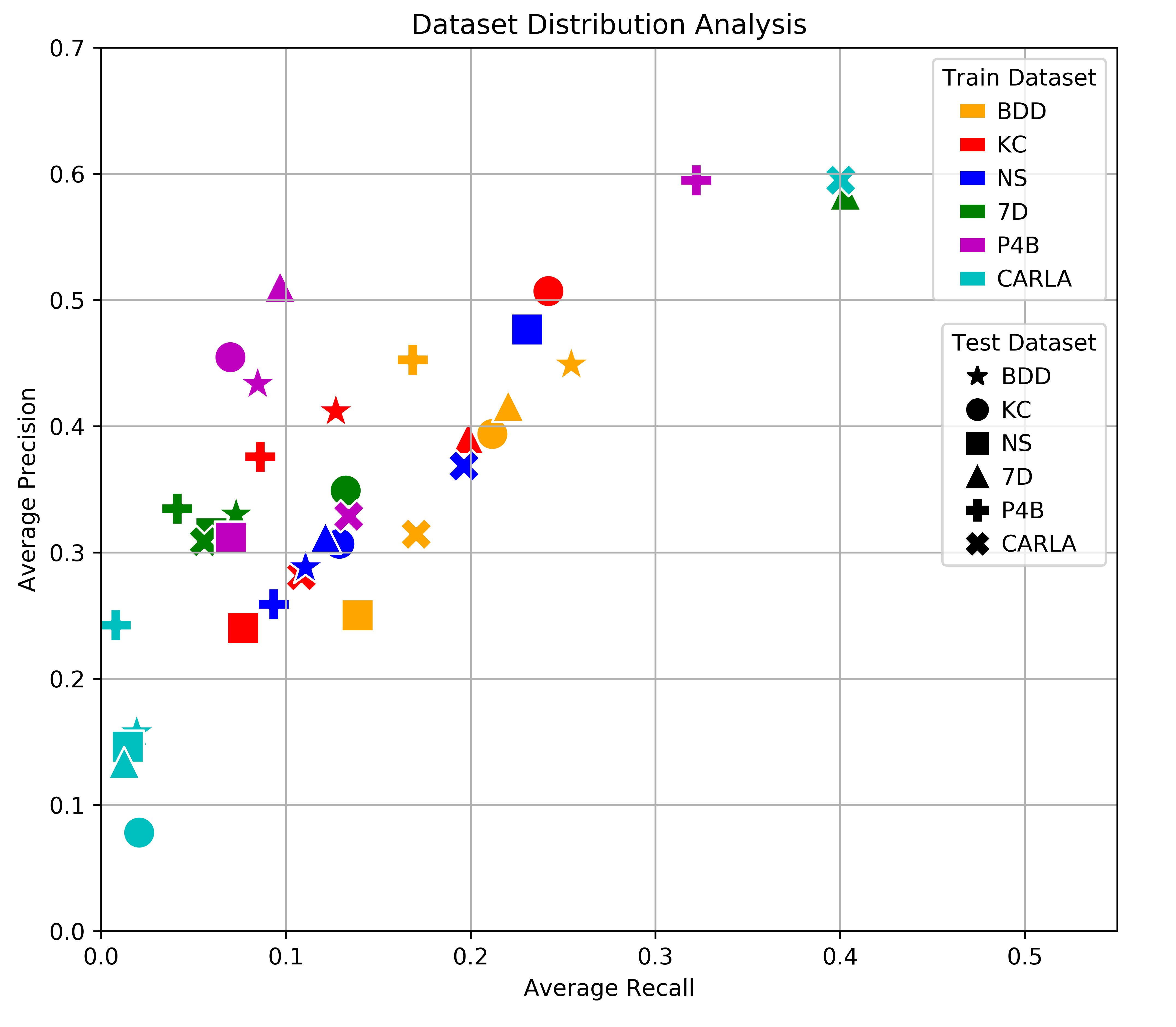}
\end{center}
\vskip -0.2in
   \caption{The model is trained on each of the datasets and is tested on all the other datasets. In the legend, the train-test dataset combinations are shown as a tuple.}
\label{fig:similarity}
\end{figure}
In our second study, we are analyzing the similarities between various datasets. This study will provide us with the knoweldge of how close the underlying distribution of one dataset is compared to the other datasets. In this section, all the datasets are trained at full training set size to achieve their best results on their own test set. Then, their trained model is used to evaluate their performance on other datasets. The results of this section are presented in Figure \ref{fig:similarity}.

Based on the distribution of the results, we can conclude some insightful information. BDD has the largest distribution in its sample space. This is visible from the more localized test results of the BDD trained model on the BDD test set and the non-BDD test sets. It is followed by KC and NS.

\begin{table}[!ht]
\caption{Results for Mixed Training. 10\%, 5\%, and 2.5\% of real data is used in a mixed training procedure with the synthetic data. Test results are reported on the test set of the corresponding test splits of the real datasets.}
\label{tab:mixedtrain}
\vskip 0.15in
\begin{center}
\begin{small}
\begin{sc}
\resizebox{.95\columnwidth}{!}{%
\begin{tabular}{lr|cc|cc}
\toprule
\multicolumn{6}{c}{\textbf{(a) 10\% Real Data}} \\
\midrule
\multicolumn{2}{c|}{\textbf{Training Dataset}}    & \multicolumn{2}{c|}{\textbf{Person}}                  & \multicolumn{2}{c}{\textbf{Car}}                    \\
\multicolumn{1}{c}{Synthetic} & \multicolumn{1}{c|}{Real} & \multicolumn{1}{c}{Precision} & \multicolumn{1}{c|}{Recall} & \multicolumn{1}{c}{Precision} & \multicolumn{1}{c}{Recall} \\
\midrule
\multirow{3}{*}{7D}
&BDD  &0.299 &0.087 &0.563 &0.320 \\
&KC   &0.394 &0.166 &0.569 &0.297 \\
&NS   &0.385 &0.060 &0.502 &0.238 \\
\midrule
\multirow{3}{*}{P4B}
&BDD  &0.284 &0.089 &0.612 &0.337 \\
&KC   &0.425 &0.116 &0.586 &0.301 \\
&NS   &0.349 &0.084 &0.525 &0.254 \\
\midrule
\multirow{3}{*}{CARLA}
&BDD  &0.228 &0.052 &0.593 &0.332 \\
&KC   &0.426 &0.060 &0.561 &0.284 \\
&NS   &0.326 &0.061 &0.514 &0.242 \\
\bottomrule
\end{tabular}
}
\end{sc}
\end{small}
\end{center}

\begin{center}
\begin{small}
\begin{sc}
\resizebox{.95\columnwidth}{!}{%
\begin{tabular}{lr|cc|cc}
\toprule
\multicolumn{6}{c}{\textbf{(b) 5\% Real Data}} \\
\midrule
\multicolumn{2}{c|}{\textbf{Training Dataset}}    & \multicolumn{2}{c|}{\textbf{Person}}                  & \multicolumn{2}{c}{\textbf{Car}}                    \\
\multicolumn{1}{c}{Synthetic} & \multicolumn{1}{c|}{Real} & \multicolumn{1}{c}{Precision} & \multicolumn{1}{c|}{Recall} & \multicolumn{1}{c}{Precision} & \multicolumn{1}{c}{Recall} \\
\midrule
\multirow{3}{*}{7D}
&BDD  &0.330 &0.077 &0.546 &0.269 \\
&KC   &0.343 &0.162 &0.530 &0.272 \\
&NS   &0.340 &0.052 &0.495 &0.204 \\
\midrule
\multirow{3}{*}{P4B}
&BDD  &0.290 &0.088 &0.566 &0.310 \\
&KC   &0.364 &0.115 &0.590 &0.253 \\
&NS   &0.343 &0.054 &0.511 &0.222 \\
\midrule
\multirow{3}{*}{CARLA}
&BDD  &0.222 &0.034 &0.570 &0.287 \\
&KC   &0.325 &0.064 &0.546 &0.260 \\
&NS   &0.349 &0.037 &0.495 &0.223 \\
\bottomrule
\end{tabular}
}
\end{sc}
\end{small}
\end{center}

\begin{center}
\begin{small}
\begin{sc}
\resizebox{.95\columnwidth}{!}{%
\begin{tabular}{lr|cc|cc}
\toprule
\multicolumn{6}{c}{\textbf{(c) 2.5\% Real Data}} \\
\midrule
\multicolumn{2}{c|}{\textbf{Training Dataset}}    & \multicolumn{2}{c|}{\textbf{Person}}                  & \multicolumn{2}{c}{\textbf{Car}}                    \\
\multicolumn{1}{c}{Synthetic} & \multicolumn{1}{c|}{Real} & \multicolumn{1}{c}{Precision} & \multicolumn{1}{c|}{Recall} & \multicolumn{1}{c}{Precision} & \multicolumn{1}{c}{Recall} \\
\midrule
\multirow{3}{*}{7D}
&BDD  &0.318 &0.069 &0.518 &0.251 \\
&KC   &0.337 &0.158 &0.531 &0.254 \\
&NS   &0.294 &0.042 &0.469 &0.180 \\
\midrule
\multirow{3}{*}{P4B}
&BDD  &0.298 &0.069 &0.533 &0.307 \\
&KC   &0.343 &0.110 &0.525 &0.261 \\
&NS   &0.267 &0.040 &0.481 &0.169 \\
\midrule
\multirow{3}{*}{CARLA}
&BDD  &0.242 &0.013 &0.515 &0.257 \\
&KC   &0.299 &0.034 &0.484 &0.228 \\
&NS   &0.271 &0.035 &0.479 &0.174 \\
\bottomrule
\end{tabular}
}
\end{sc}
\end{small}
\end{center}
\vskip -0.1in
\end{table}

All the synthetic datasets suffer from a specificity problem which results in models that are incapable of proper generalization. They perform very well on their own test set, however their performance suffers on any other test sets. We can see P4B is a better training dataset than 7D, despite the photo-realistic nature of 7D. CARLA has less accurate environment and camera models compared to other synthetic datasets that results in its poor performance.

\begin{table}[!ht]
\caption{Results for Fine-tuning with Real Data. Model is trained on the synthetic dataset and is then fine-tuned on a 10\%, 5\%, and 2.5\% portion of the real dataset. Test results are reported on the test set of the corresponding test splits of the real datasets.}
\label{tab:finetune}
\vskip 0.15in
\begin{center}
\begin{small}
\begin{sc}
\resizebox{.95\columnwidth}{!}{%
\begin{tabular}{lr|cc|cc}
\toprule
\multicolumn{6}{c}{\textbf{(a) 10\% Real Data}} \\
\midrule
\multicolumn{2}{c|}{\textbf{Training Dataset}}    & \multicolumn{2}{c|}{\textbf{Person}}                  & \multicolumn{2}{c}{\textbf{Car}}                    \\
\multicolumn{1}{c}{Synthetic} & \multicolumn{1}{c|}{Real} & \multicolumn{1}{c}{Precision} & \multicolumn{1}{c|}{Recall} & \multicolumn{1}{c}{Precision} & \multicolumn{1}{c}{Recall} \\
\midrule
\multirow{3}{*}{7D}
&BDD  &0.280 &0.139 &0.578 &0.435 \\
&KC   &0.349 &0.179 &0.609 &0.375 \\
&NS   &0.364 &0.129 &0.532 &0.341 \\
\midrule
\multirow{3}{*}{P4B}
&BDD  &0.298 &0.114 &0.594 &0.404 \\
&KC   &0.351 &0.154 &0.595 &0.373 \\
&NS   &0.330 &0.118 &0.522 &0.328 \\
\midrule
\multirow{3}{*}{CARLA}
&BDD  &0.213 &0.077 &0.565 &0.394 \\
&KC   &0.339 &0.117 &0.598 &0.336 \\
&NS   &0.297 &0.096 &0.509 &0.306 \\
\bottomrule
\end{tabular}
}
\end{sc}
\end{small}
\end{center}

\begin{center}
\begin{small}
\begin{sc}
\resizebox{.95\columnwidth}{!}{%
\begin{tabular}{lr|cc|cc}
\toprule
\multicolumn{6}{c}{\textbf{(b) 5\% Real Data}} \\
\midrule
\multicolumn{2}{c|}{\textbf{Training Dataset}}    & \multicolumn{2}{c|}{\textbf{Person}}                  & \multicolumn{2}{c}{\textbf{Car}}                    \\
\multicolumn{1}{c}{Synthetic} & \multicolumn{1}{c|}{Real} & \multicolumn{1}{c}{Precision} & \multicolumn{1}{c|}{Recall} & \multicolumn{1}{c}{Precision} & \multicolumn{1}{c}{Recall} \\
\midrule
\multirow{3}{*}{7D}
&BDD  &0.257 &0.118 &0.581 &0.402 \\
&KC   &0.348 &0.161 &0.557 &0.369 \\
&NS   &0.312 &0.125 &0.498 &0.317 \\
\midrule
\multirow{3}{*}{P4B}
&BDD  &0.323 &0.071 &0.585 &0.373 \\
&KC   &0.312 &0.145 &0.567 &0.360 \\
&NS   &0.358 &0.092 &0.552 &0.253 \\
\midrule
\multirow{3}{*}{CARLA}
&BDD  &0.170 &0.066 &0.559 &0.369 \\
&KC   &0.334 &0.091 &0.575 &0.337 \\
&NS   &0.303 &0.077 &0.487 &0.298 \\
\bottomrule
\end{tabular}
}
\end{sc}
\end{small}
\end{center}

\begin{center}
\begin{small}
\begin{sc}
\resizebox{.95\columnwidth}{!}{%
\begin{tabular}{lr|cc|cc}
\toprule
\multicolumn{6}{c}{\textbf{(c) 2.5\% Real Data}} \\
\midrule
\multicolumn{2}{c|}{\textbf{Training Dataset}}    & \multicolumn{2}{c|}{\textbf{Person}}                  & \multicolumn{2}{c}{\textbf{Car}}                    \\
\multicolumn{1}{c}{Synthetic} & \multicolumn{1}{c|}{Real} & \multicolumn{1}{c}{Precision} & \multicolumn{1}{c|}{Recall} & \multicolumn{1}{c}{Precision} & \multicolumn{1}{c}{Recall} \\
\midrule
\multirow{3}{*}{7D}
&BDD  &0.262 &0.096 &0.561 &0.335 \\
&KC   &0.337 &0.154 &0.565 &0.355 \\
&NS   &0.288 &0.107 &0.463 &0.316 \\
\midrule
\multirow{3}{*}{P4B}
&BDD  &0.329 &0.059 &0.597 &0.299 \\
&KC   &0.329 &0.118 &0.556 &0.333 \\
&NS   &0.363 &0.071 &0.543 &0.210 \\
\midrule
\multirow{3}{*}{CARLA}
&BDD  &0.167 &0.043 &0.549 &0.335 \\
&KC   &0.281 &0.085 &0.560 &0.317 \\
&NS   &0.250 &0.071 &0.449 &0.282 \\
\bottomrule
\end{tabular}
}
\end{sc}
\end{small}
\end{center}
\vskip -0.1in
\end{table}

Performance of all the real data trained models on the 7D test set is much better than other synthetic test sets. This means that 7D has a nature that is better covered by the real dataset. This is also confirmed in \cite{7D}.

\subsection{Synthetic-Real Data Mixing}
\begin{figure}[!ht]
\begin{center}
   \includegraphics[width=0.99\linewidth]{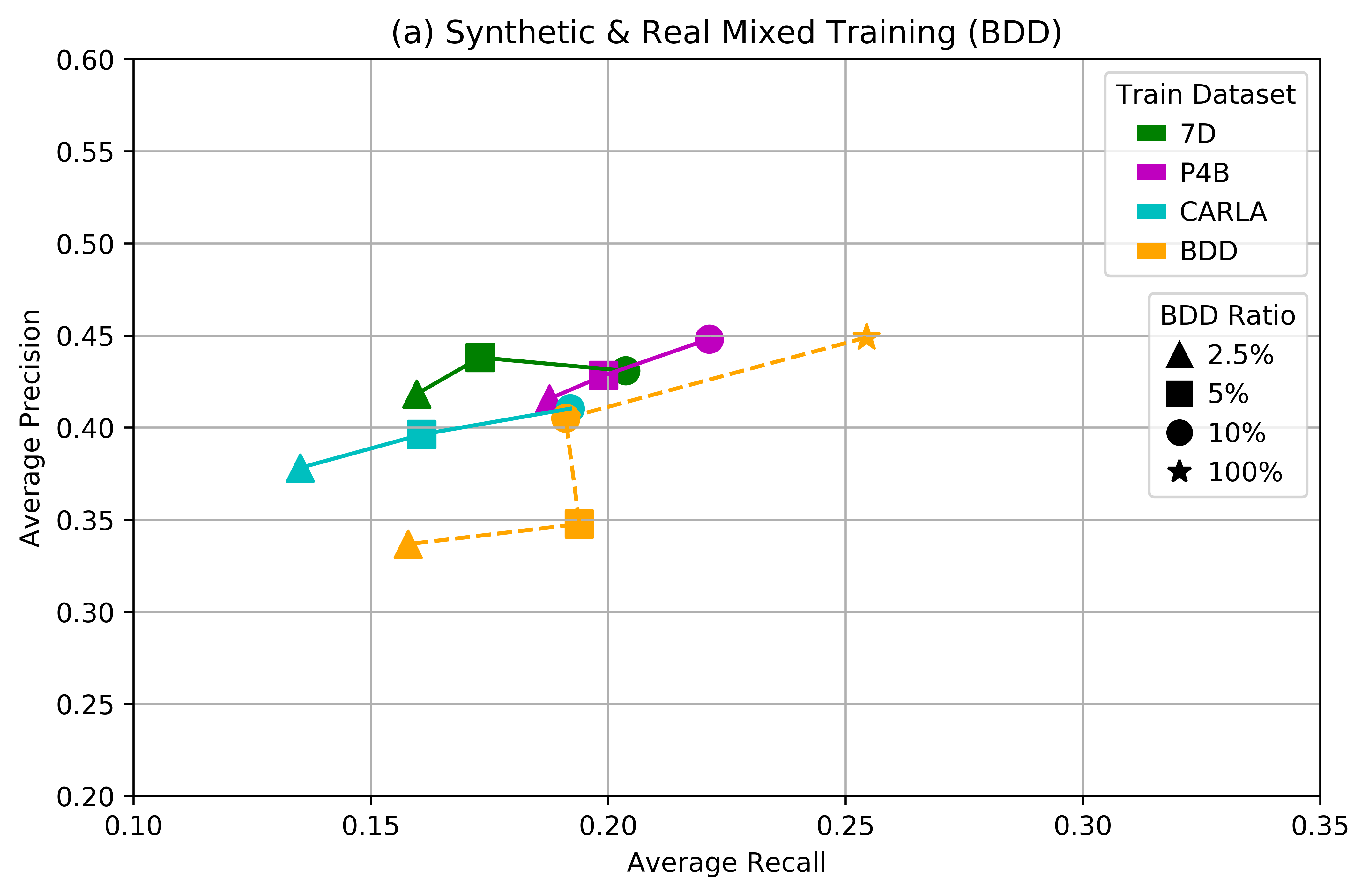}
   \includegraphics[width=0.99\linewidth]{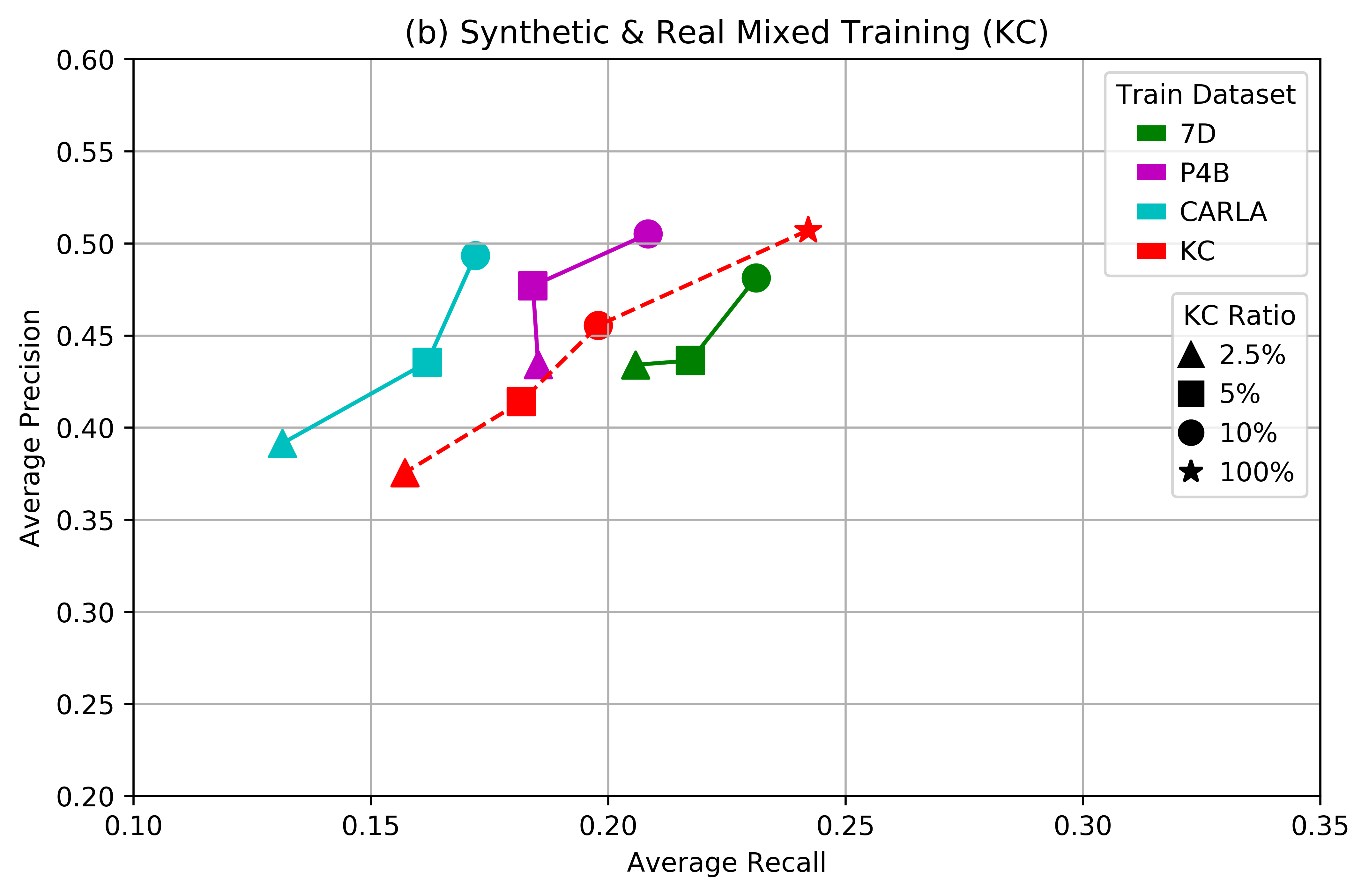}
   \includegraphics[width=0.99\linewidth]{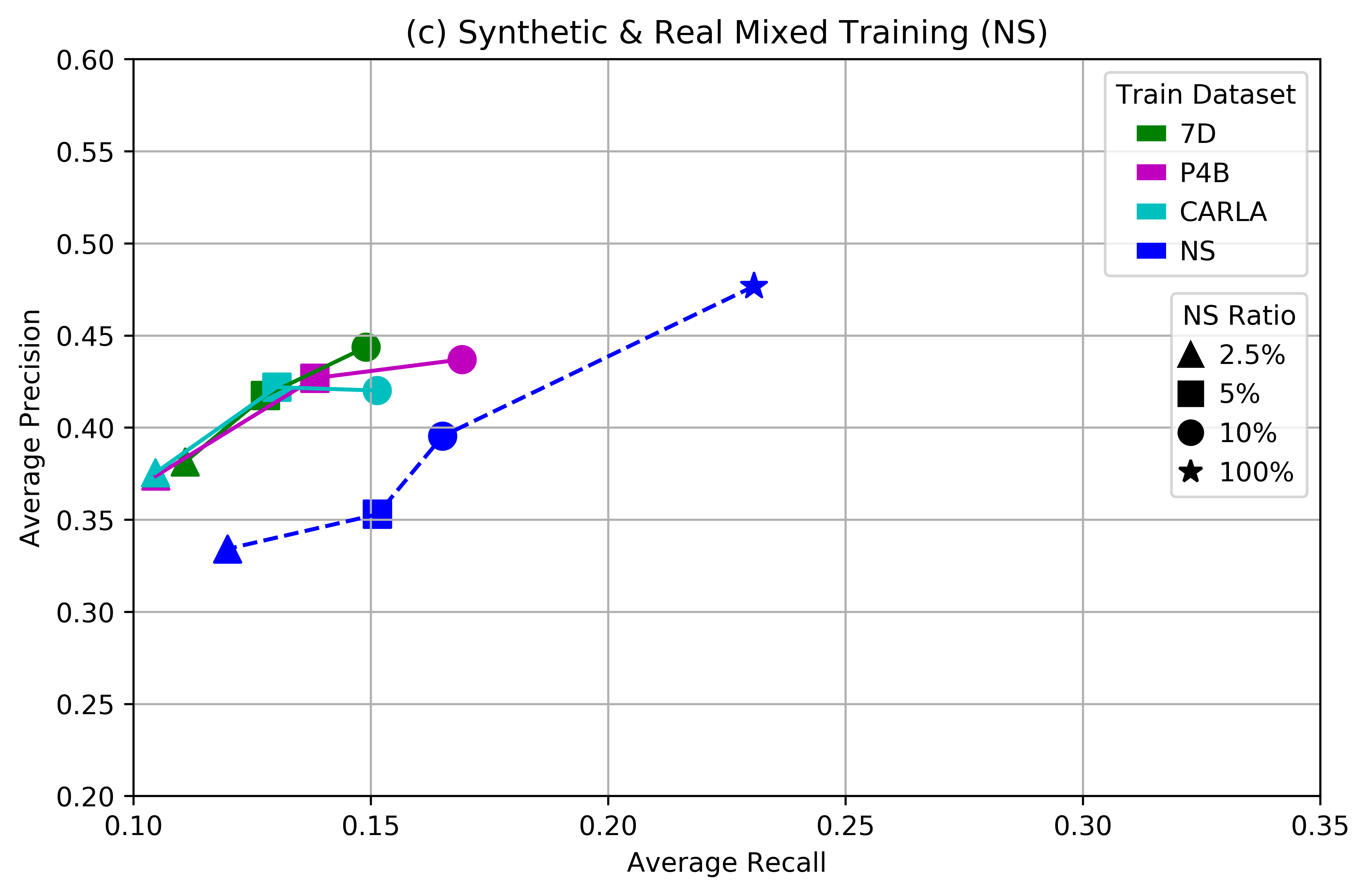}
\end{center}
   \caption{Results for Mixed Training.  Using mixed (synthetic and real) dataset for training. The test are performed on the test split of the real dataset.}
\label{fig:mix}
\end{figure}

\begin{figure}[!ht]
\begin{center}
   \includegraphics[width=0.99\linewidth]{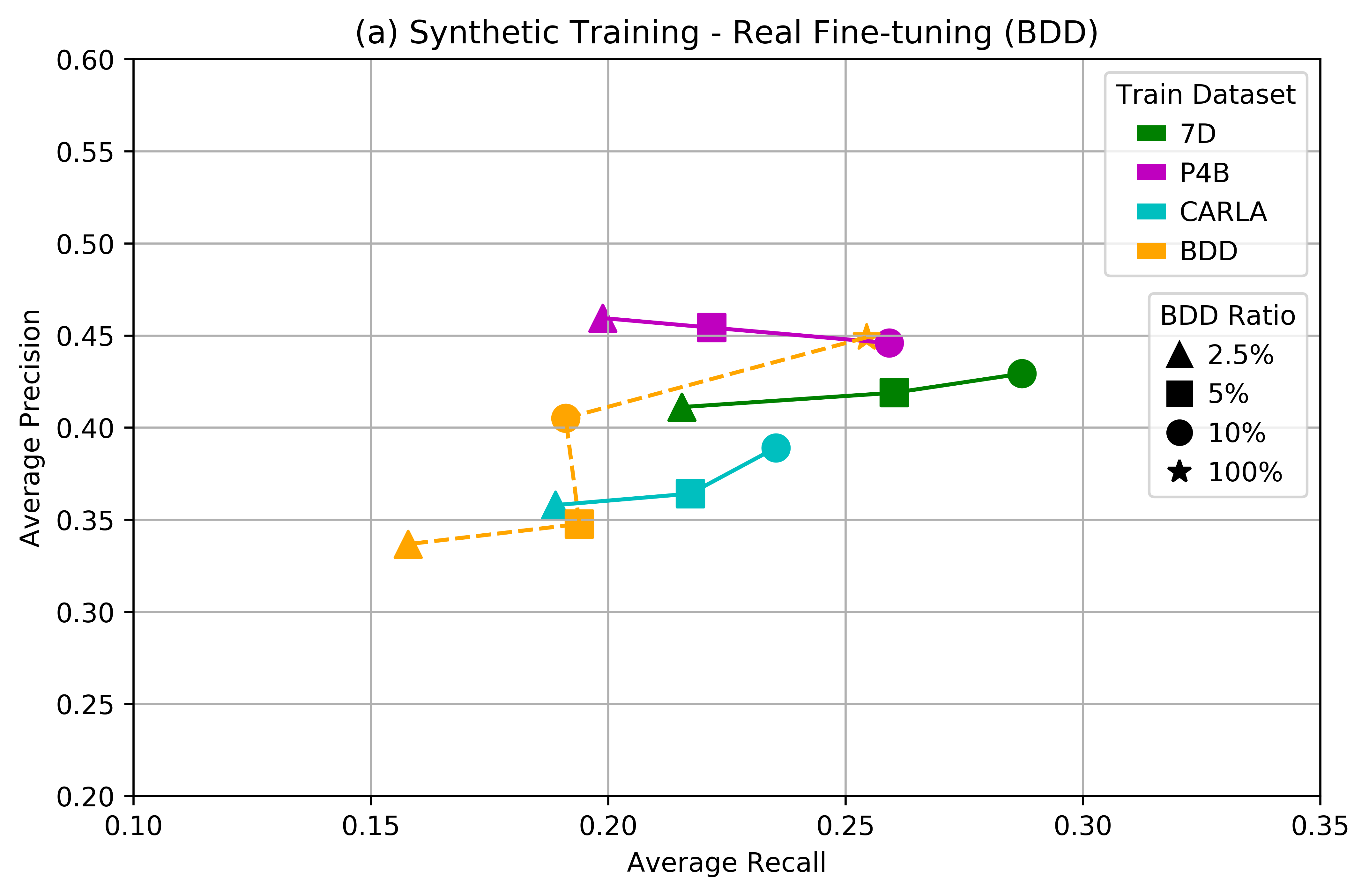}
   \includegraphics[width=0.99\linewidth]{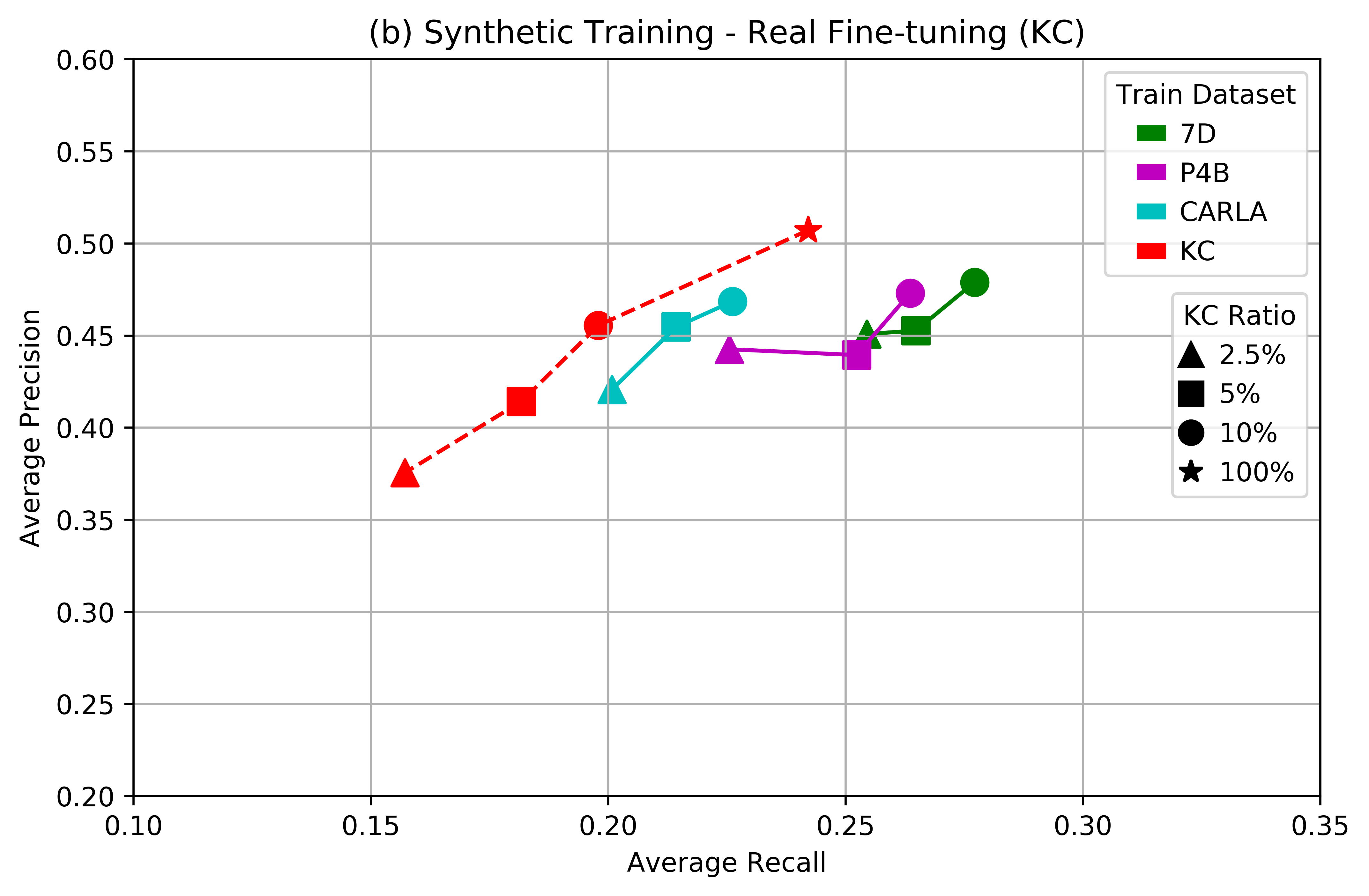}
   \includegraphics[width=0.99\linewidth]{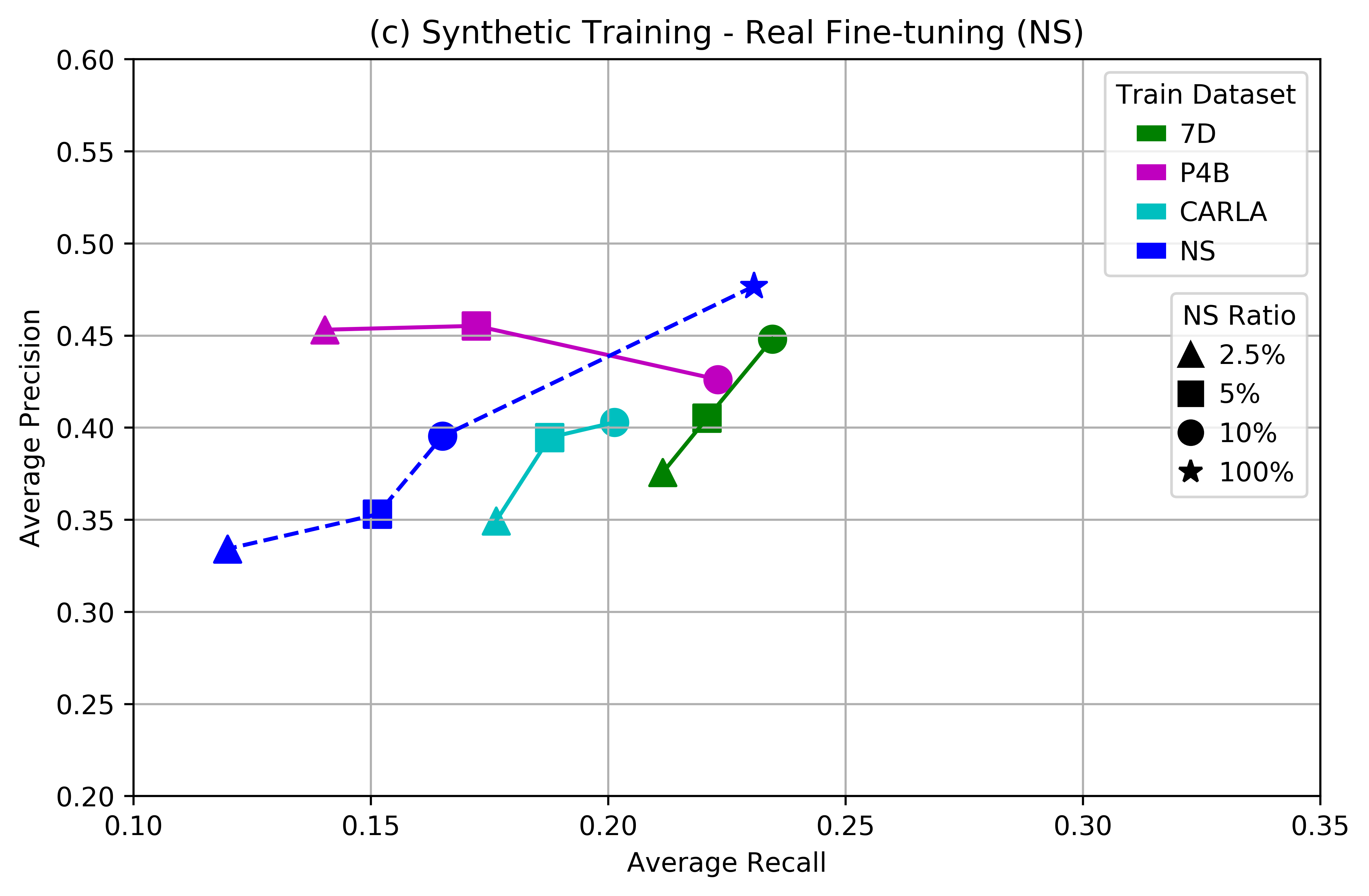}
\end{center}
   \caption{Results of Fine-tuning with Real Data. Model is trained on the synthetic dataset and is then fine-tuned on a real dataset. The test results are performed on the test split of the real dataset.}
\label{fig:finetune}
\end{figure}

In an attempt to achieve the full real dataset performance with only using a fraction of it, we launch a study that uses a mixed set of synthetic and real datasets with various ratios. These per-class results are shown in table \ref{tab:ratios}, while the averaged results are shown in figure \ref{fig:mix}. The mixed dataset is used in training, while the tests are performed only on the  corresponding test set used real dataset.

Table \ref{tab:mixedtrain} shows the precision and recall rates of the mixed datasets on a per-class basis. These results are directly comparable to the results achieved by only using a fraction of the real data in table \ref{tab:ratios}. Using both P4B and KC in mixed training provides valuable diversity of features that results in surpassing full real dataset training performance of 'person' detection while only using 10\% of real data. We cannot see the same performance in case of the 'car' class as it conveys lesser varying structure compared to the 'person' class. However, the performance of the 'car' class is better than only using 10\% of real data.

In this section, our hypothesis to observe a performance increase by adding a small amount of real data compared to the synthetic only training is confirmed. This implies that a rather huge amount of cost saving could be achieved in data annotation.

We can further see that increasing the real dataset ratio from 2.5\% to 5\% and 10\% in all cases results in an increase in the performance. These gains could be observed as an increase in precision, recall, or both of the measures.

On average P4B is providing better results on BDD and NS. CARLA is consistently inferior to the other sophisticated synthetic datasets.
\subsection{Synthetic Training and Real Data Fine-tuning}
In mixed training, our model learns the general concepts from simulated datasets, and uses the real samples to adapt its domain. However, there is no scheduling in the mixed training sessions. To perform a more structured experiment, we take a transfer learning approach. Model is first trained on a synthetic dataset, and then fine-tuned on each of the real datasets. We use the same ratios defined in the previous section. Similarly, the tests take place on the real data only. The results of this section are presented in Table \ref{tab:finetune} and in figure \ref{fig:finetune}.

Employing transfer learning strategy significantly increases the recall rates for all of the models. This is attributed to the fact that the model is not only able to transfer knowledge from synthetic data, but is also capable of expanding its learned features with real data to achieve a better generalization. 7D is specially benefiting from the fine-tuning in comparison to mixed training. The 'person' class in particular enjoys a dramatic increase in recall for all the experiments. The smaller variation in the results of each experiment shows that transfer learning is a more stable approach for training deep models.

Using this methodology, the exhaustive training and faster fine-tuning phases can be separated from each other, and the same base model could be easily adapted to multiple domains. This is highly valuable as it enables the base model to be trained on a complete synthetic dataset in advance. Later, it could be quickly fine-tuned on a geography specific dataset and deployed on vehicles of that region.

\subsection{Combination of Synthetic Datasets}
In previous tests, we showed the performance of the individual simulated datasets on each of the real datasets. Choosing the best performing synthetic dataset requires an exhaustive training and testing sessions. To avoid this complication and to address the data completeness challenge, we evaluate the effectiveness of combining multiple synthetic datasets. To perform this experiment, we combine all the synthetic datasets to train the model. Later, the model is fine-tuned on 1500 real images.

\begin{figure}[!ht]
\vskip 0.2in
\begin{center}
   \includegraphics[width=0.99\linewidth]{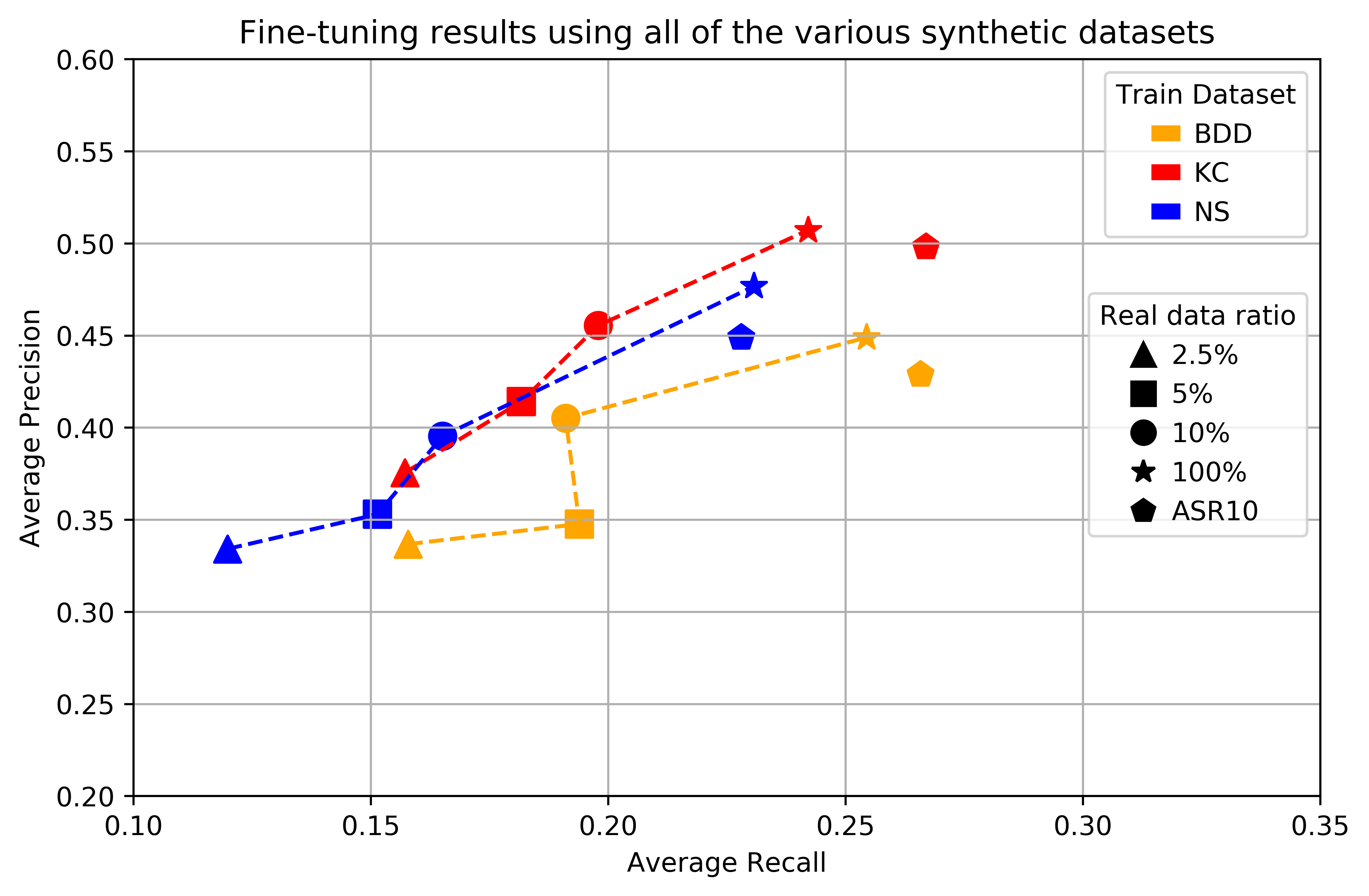}
\end{center}
\vskip -0.2in
   \caption{Results of using all the synthetic datasets together for training, and 10\% of real dataset size (3\% of train size) for fine-tuning. These results are shown as ASR10 in the figure.}
\label{fig:allsallr}
\end{figure}

Figure \ref{fig:allsallr} shows the experimental results of this task. We can see that combining all the synthetic datasets for training provides better results than individually using them, except in case of BDD. This could be attributed to the completeness of the combined training data. As each dataset is generated from an independent source, union of them presents a more complete dataset that is essential for the autonomous driving task \cite{rao2018deep}.

\section{Conclusion}
In this paper, we have cross-compared the performance of multiple datasets using SSD-MobileNet architecture in car and person detection. We have extensively analyzed the effects of training using datasets with a large amount of synthetic data and a small number of real data in two folds; mixed training and fine-tuning. Fine-tuning synthetic training model with limited real data provides better results than mixed training. This is also ultimately valuable as the huge training session can be achieved independently from the smaller fine-tuning counterpart. It is shown that the photo-realism is not as important as the diversity of the data. Objects with higher degrees of deformability require more information compared to others.

The impressive results of synthetic training are valuable, as real data is very expensive to annotate. Using simulated data as a cheaper source of training samples can provide significant savings of both cost and time. For future work, there are multiple avenues left to be explored. We would like to further expand this study to cover the meta learning models. 
Few shot learning techniques could expand this study, as they focus on generalizing a network's feature maps while using very little additional data.
Another avenue would be to study the effectiveness of various domain adaptation techniques, where a model is used to convert synthetic images to a domain more similar to the target domain, on which we then perform the training. We expect CARLA to specially benefit from this case.

We have evaluated the performance of combining multiple independent synthetic datasets. However, there needs to be additional studies on the approaches that could achieve a complete dataset using a single source while addressing all the remaining challenges related to dataset creation.

We hope this paper provides an insight on the principal dynamics between real and synthetic data, and direct future studies towards the production of cost effective procedural methodologies for training neural networks using smaller amounts of real data.


\bibliography{ICML2019_SynVsReal}
\bibliographystyle{icml2019}

%
%
%

\end{document}